\documentclass{article}
\usepackage{atbegshi}
\AtBeginDocument{\AtBeginShipoutNext{\AtBeginShipoutDiscard}}

\usepackage{graphicx}
\usepackage{amsthm}
\usepackage{amsmath}
\usepackage{amssymb}
\usepackage{array}
\usepackage[T1]{fontenc}
\usepackage{tabularx}
\usepackage{enumerate}
\usepackage{cite}
\usepackage[]{algorithm2e}
\usepackage{tabularx}
\usepackage{authblk}
\usepackage{chngpage}

\begin{document}

\title{GELATO and SAGE: An Integrated Framework for MS Annotation}
\author[1]{Khalifeh AlJadda}
\author[2]{Rene Ranzinger},
\author[2]{Melody Porterfield}
\author[3]{Mohammed Korayem}
\author[1]{John A. Miller}
\author[1]{Khaled M. Rasheed}
\author[1]{Krys J. Kochut}
\author[2]{William S. York}
\affil[1]{Department of Computer Science, University of Georgia}
\affil[2]{Complex Carbohydrate Research Center, University of Georgia}
\affil[3]{School of Informatics and Computing, Indiana University}
\maketitle

\begin{abstract}
	\textbf{Motivation:} Several algorithms and tools have been developed to (semi) automate the process of glycan identification by interpreting Mass Spectrometric data. However, each has limitations when annotating MS$^n$ data with thousands of MS spectra using uncurated public databases. Moreover, the existing tools are not designed to manage MS$^n$ data where $n > 2$.

	\textbf{Results:} Here, we propose a novel software package to automate the annotation of tandem MS data. This software consists of two major components. The first, is a free, semi-automated MS$^n$ data interpreter called the Glycomic Elucidation and Annotation Tool (GELATO). This tool extends and automates the functionality of existing open source projects, namely, GlycoWorkbench (GWB) and GlycomeDB.  The second is a machine learning model called Smart Anotation Enhancement Graph (SAGE), which learns the behavior of glycoanalysts to select annotations generated by GELATO that emulate human interpretation of the spectra.

	\textbf{Availability:} GELATO is available within GRITS-Toolbox:\\ http://www.grits-toolbox.org/
	\\SAGE to be available soon.

\end{abstract}

\section{Introduction}

Along with nucleic acids, proteins, and lipids, complex carbohydrates, also known as glycans, comprise the four major classes of macromolecules fundamental to all living systems  \cite{werz2007exploring}. Until recently, relatively little attention has been paid to studying glycans despite the major roles they play in diverse biological processes  \cite{dwek1996glycobiology}. The emergence of the field of glycobiology is warranted by the accumulated evidence for the role of glycans in cell growth and metastasis, cell-cell communication, and microbial pathogenesis. Almost all cells are coated with a dense layer of glycans and virtually all multicellular interactions take place in the context of this layer. Most proteins that are produced by eukaryotic cells for export or insertion into the cell membrane are glycosylated and proper glycosylation is often critical for their biological functions \cite{apweiler1999frequency, brooks2009strategies}. Due to their complex structure, the potential information content encoded by glycans attached to proteins exceeds that of any other post-translational modification \cite{rademacher2012glycan}. Comprehensive characterization of the glycans on glycoproteins has become an essential element for drug development, quality control, and basic biomedical research. However, glycan identification is much more difficult than protein identification, and de novo glycan sequencing is a proven NP-hard problem \cite{tang2005automated}. Glycans are more diverse than nucleic acids and proteins and peptides \cite{goldberg2006automatic,werz2007exploring}, mainly due to their branched structures.  The linear structures of peptides and the availability of reliable peptide sequence databases facilitate their identification by tandem mass spectrometry (MS$^n$) \cite{aebersold2003mass} which generates a relatively complete series of high-intensity fragment ions with mass differences that correspond to specific amino acids and thus provides clear-cut information regarding the peptide’s amino acid sequence. In contrast, the branched nature of glycans often precludes the generation of readily interpretable ion series during the MS$^n$ analysis . Furthermore, glycans are composed of monosaccharide building blocks that comprise isomeric sets (e.g., the set of all hexoses) whose members can be rarely distinguished by MS.  Finally, annotation of MS data with structures from glycan databases is error-prone because existing glycan databases are incomplete, minimally curated, and frequently polluted with erroneous or irrelevant structures.  

Current glycomics technology thus relies heavily on the manual interpretation of mass spectrometry datasets. As instrumentation improves, dataset size increases (e.g., 2000 or more mass spectra per biological sample), thus demanding more time for manual interpretation and reducing the number of samples that can be analyzed. This bottleneck is a major impediment blocking the application of glycoanalysis to a broad range of important biomedical investigations. Our evaluation of the existing tools for MS annotation revealed many limitations (Table \ref{tbl:comp}).  GlycoMod \cite{cooper2001glycomod} is a free web-based tool intended for MS profile annotation but not for MS$^n$ annotation.  It annotates spectral features with compositions rather than structures. This tool supports limited types of chemical derivitisation and ionization adducts, with no support for neutral exchange. 
GlycoPeakfinder \cite{maass2007glyco} is another web based tool that annotates both MS profiles and MS$^n$ spectra. However, it only allows annotation of one spectrum at a time, and annotates with compositions rather than structures. GlycoWorkbench (GWB) \cite{ceroni2008glycoworkbench} is freely-available software that annotates MS$^n$ spectra using user defined structures or structures from a database. Its main limitation is that it can upload and process only one spectrum at a time, slowing down the data processing steps. Furthermore, incomplete chemical methylation of glycan structures \cite{cheng2013increasing} is not considered in the annotation process. The commercial tool SimGlycan$^{\textregistered}$ provides annotation of MS$^n$ spectra with glycan structures from an integrated database based on KEGG \cite{kanehisa2002kegg}. It facilitates high throughput analysis by uploading and annotating entire MS$^n$ runs, but does not consider undermethylation of glycans. Here, we describe a new software package to facilitate the interpretation of MS$^n$ data.  This package consists of two components: GELATO, a freely available algorithm for the annotation of MS$^n$ spectra of glycans, and SAGE, a machine learning model for refining GELATO annotations with trained expert knowledge.

\section{Methods}
	
	Glycomic Elucidation and Annotation Tool (GELATO) is a freely available, semi-automated interpreter for MS$^n$ of glycans designed and implemented at the Complex Carbohydrate Research Center (CCRC). GELATO extends the functionality of existing open source projects, namely, GlycoWorkbench (GWB) \cite{ceroni2008glycoworkbench} and GlycomeDB \cite{ranzinger2011glycomedb}.  
	
	The following extensions, which are not part of the GWB annotation algorithm, are implemented in GELATO:
	\begin{enumerate}
		\item Uploading the complete data set from an MS$^n$ run at once rather than requiring each spectrum to be uploaded separately.
		\item Specification of ionization adducts not predefined in our list on the basis of user-supplied information, including name, charge, and mass.
		\item Annotation of multiply charged ions.
		\item Annotation of ionic species generated by neutral ion exchange processes (e.g., replacement of H$^+$ with Na$^+$).
		\item Annotation of ions generated by loss of small molecules (e.g., water or methanol).
		\item Identification of ions arising from incompletely methylated glycans.
		\item Ability to account for different fragmentation processes depending on the MS level or ion-fragmentation method.
		\item Support for different accuracy (mass tolerance) settings for MS$^1$ and MS$^n$, consistent with various spectrometer setup protocols.
	\end{enumerate}
	
\begin{figure}[!tpb]
		\centerline{\includegraphics[scale=0.5]{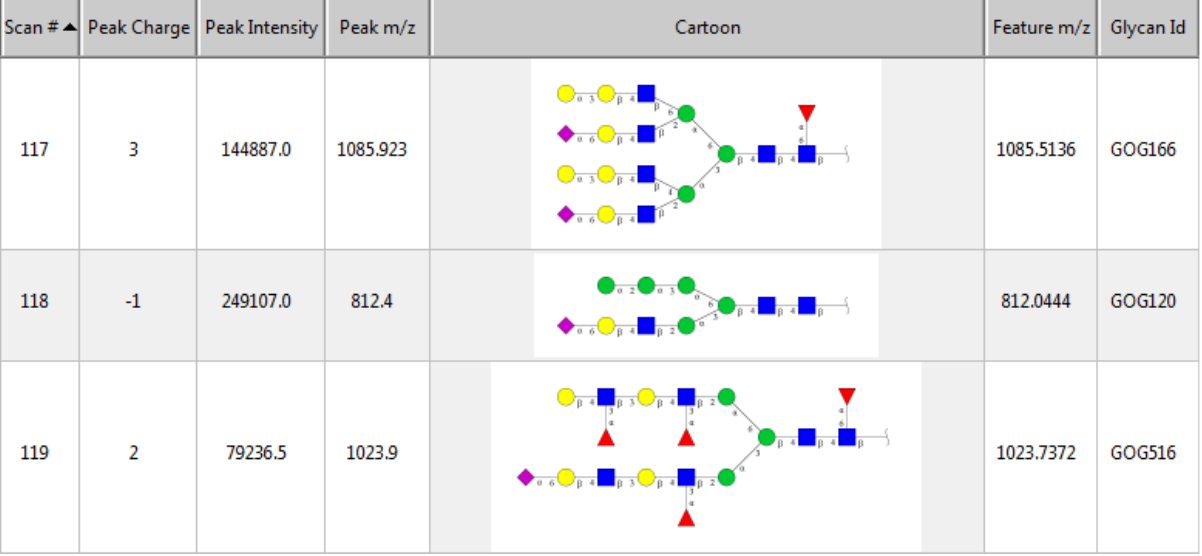}}
		\caption{MS$^1$ annotations using GELATO}\label{fig:08}
	\end{figure}

\begin{figure}[!tpb]
		\centerline{\includegraphics[scale=0.5]{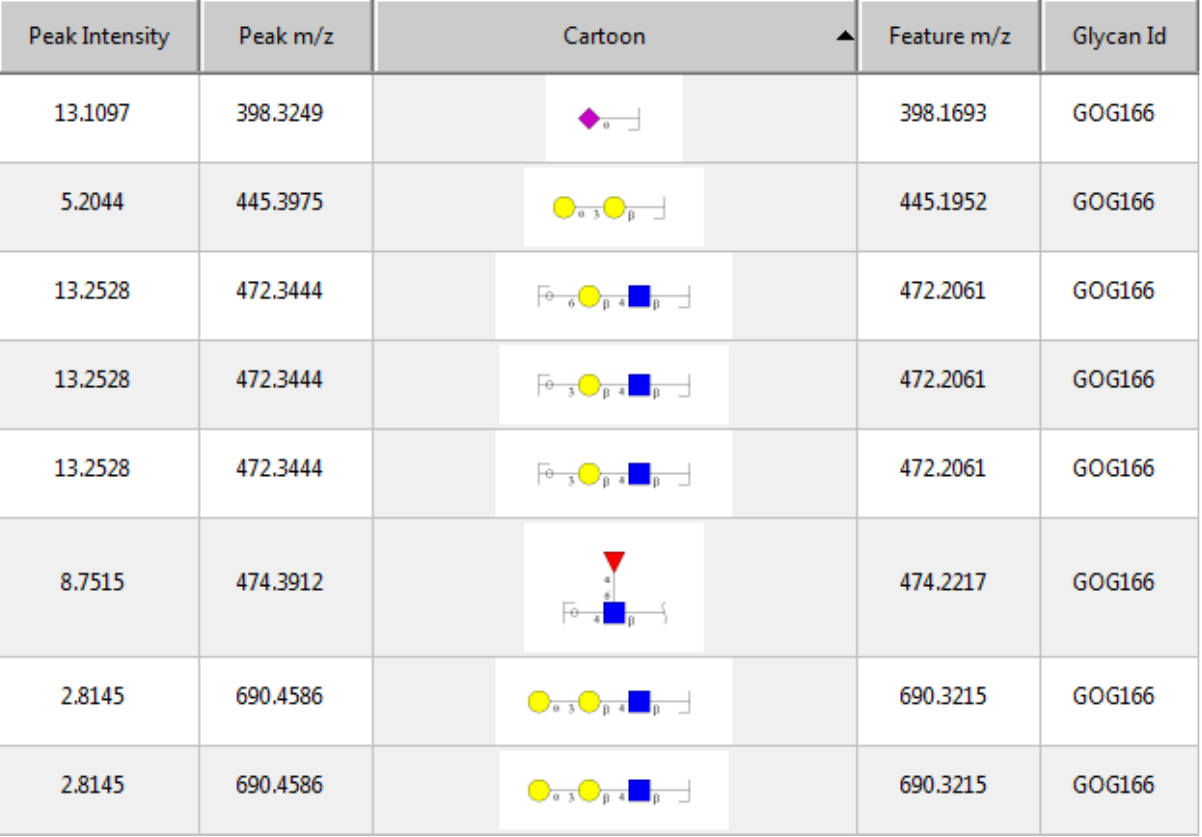}}
		\caption{MS$^n$ annotations using GELATO}\label{fig:09}
	\end{figure}

	Confidence in the results produced by GELATO is increased by using a set of human curated databases generated from the glycan structure ontology, GlycO \cite{thomas2006modular}, as the source of glycan structures for the annotations of different types of glycans (e.g, N-glycans, O-glycans and Glycosphintolipids) \cite{eavenson2014qrator}. Figure \ref{fig:08} and Figure \ref{fig:09} show the annotations produced by GELATO for MS$^1$ and MS$^n$, respectively. (The annotation workflow used by GELATO is provided in the supplementary document.) The GELATO annotation process starts by setting parameters that specify the types of ions (e.g., Na+ adducts), neutral ion exchange events, and fragmentation processes (including, e.g., glycosidic bond cleavage and loss of water or methanol) that the user deems likely to generate the spectra being processed. Candidate glycan structures are then retrieved from the chosen database one at a time and each is checked to determine whether its calculated mass corresponds, within a specified tolerance, to the $m/z$ of any precursor ion detected in MS$^1$. If a match is found, fragmentation of the candidate glycan is simulated in silico using the user-specified settings, and the calculated $m/z$ values and structure of each theoretical fragment ion is saved.,The $m/z$ values of the simulated fragment ions are compared with the $m/z$ values of ions observed in the spectrum generated by fragmentation of the matching precursor ion, allowing observed (precursor and fragment)  ions to be annotated with theoretical candidate structures. The resulting annotations are immediately serialized to a data file as they are assigned, making it possible for GELATO to run on a desktop computer or laptop with limited memory and CPU speed to handle MS data files containing hundreds of thousands of spectra. For MS$^n$ spectra, this process is repeated recursively by choosing appropriate fragment ion annotations as candidate precursor structures for further $in silico$ fragmentation and annotation of the spectra at the next MS level. GELATO generates and records all of the possible annotations based on these criteria for each processed spectrum without applying human expert knowledge or filtering the results. The annotations (each corresponding to a different candidate glycan structure) are then ranked using two complementary scoring metrics. The first, $score_c$, is calculated by dividing the number of fragment annotations for glycan $G_x$ by the total number of ions in scan $S_y$, as shown in Equation \ref{eq:01} (Figure \ref{fig:eq}). Although the structurally relevant ions in a given spectrum usually exhibit high intensity, $score_c$ incorporates all annotated peaks regardless of their intensity. This can lead to less meaningful rankings if a large number of noise peaks are annotated because they happen to have $m/z$ values that match the simulated fragments of the candidate glycan. This issue is addressed by calculating a second metric ($score_i$), which corresponds to the fraction of the total observed ion intensity in a fragment-ion spectrum that is derived from annotated peaks (Equation  \ref{eq:02} in Figure \ref{fig:eq}). However, $score_i$ may also result in a misleading ranking if only one or few high intensity peaks are annotated. Therefore, both scores are provided to assist in user evaluation of the annotation results.

	\begin{figure}
		\begin{equation}
		score_c(G_x|S_y) = \frac{num(AP_{xy})}{num(P_y)}\label{eq:01}
		\end{equation}
		\begin{equation}
		score_i(G_x|S_y) = \frac{\sum I(AP_{xy})}{\sum I(P_y)}\label{eq:02}
		\end{equation}
		\caption{Scoring metrics calculated by GELATO.  The variable num($AP_{xy}$) is the number of peaks in spectrum $S_y$ that are annotated by simulated fragments of $G_x$ while $num(P_y)$ is the number of peaks in spectrum $S_y$. $I(AP_{xy})$ is the total intensity of peaks in spectrum $S_y$ that are annotated by simulated fragments of $G_x$ while $I(P_y)$ is the total intensity of peaks in spectrum $S_y$.\label{fig:eq}}
	\end{figure}
	
	TThe GELATO annotation generation and ranking processes provide all theoretically possible annotations (for structures in the database), many of which may be meaningless. The user must therefore review the annotations and eliminate those that are judged to be incorrect, which can take considerable time and effort, especially when high throughput data are processed. In order to speed up the annotation of MS data, we thus implemented a machine learning model that identifies meaningful annotations based on human review patterns of previously processed datasets. This algorithm was implemented as a customization of the Probabilistic Graphical Model for Massive Hierarchical Data (PGMHD) \cite{aljadda2014pgmhd}. The aim of the resulting tool, which we call Smart Annotation Enhancement Graph (SAGE), is to learn the annotation behavior of a user or group of users and apply this annotation behavior in subsequent annotations. The tool builds a probabilistic graph model that represents glycans and glycan fragments previously selected as meaningful annotations by the user and utilizes this graph to calculate the probability that the user would accept or reject a given annotation of the new dataset. SAGE can be used in two contexts, as an annotation tool, where it actually generates, scores and selects annotations, or as post-filtering tool, where it analyzes previously annotated spectra (e.g, processed by GELATO) to reject annotations that are unlikely to be accepted by the user. Training SAGE is straightforward and can be accomplished either in a single session or over many sessions. For example, spectra that have been annotated using GELATO might be reviewed by the user who selects a subset of the provided annotations that he/she judges to be correct. The selected annotations cann then be processed by SAGE to either build a new probabilistic graph model or integrate the new annotations into an existing model. The proposed learning algorithm for SAGE, shown in Algorithm \ref{alg:learn}, is designed to facilitate progressive learning, which has the following advantages:

	\begin{enumerate}
		\item Data required for training the model can be generated, evaluated by the user and processed in stages and at different times.
		\item New training data is easily incorporated to extend the model without reprocessing data used in previous training sessions.
		\item Training can be distributed over many sessions, eliminating the need for a single, prolonged session, which might fail and have to be repeated.
		\item Recursive learning is possible, allowing the model itself to generate new training data by processing new MS data sets, provided that the new annotations are judged to be accurate by the user.
	\end{enumerate}  
	
	\begin{algorithm}
		\label{alg:learn}
		\KwData{Annotated MS$^n$ Spectra Using GELATO}
		\KwResult{SAGE Instance}
		
		\Begin{
			$currentMSLevel = 0$\\
			\While {$currentMSLevel < maxMSLevel-1$}{
				\ForEach {$annotatedPrecursor \in currentMSLevel$}{
					\eIf {$annotatedPrecursor.annotation$ exists in $SAGE.currentLevelNodes$}{
						get $sageNode$ where $sageNode.annotation$ = $annotatedPrecursor.annotation$\\
						$sageNode.frequency += 1$
					}{
					$sageNode = new node$\\
					$sageNode.frequency = 1$
				}
				$childrenLevel = currentMSLevel+1$\\
				\ForEach {$annotatedChildPeak \in annotatedPrecursor.peakList$}{
					\ForEach {$sageChildNode \in sageNode.children$}{
						\eIf {$annotatedChildPeak.annotation = sageChildNode.annotation$}{
							$edge = edge(sagedNode,sageChildNode)$
							$edge.frequency += 1$
						}{\eIf{$childNode \in sage.childrenLevelNodes$}{
							$edge = createNewEdge(sageNode,sageChildNode)$
							$edge.frequency = 1$
						}{
						$sageChildNode = new Node$
						$sageChildNode.annotation = annotatedChildPeak.annotation$
						$sageChildNode.frequency = 1$
						$edge = createNewEdge(sageNode,sageChildNode)$
						$edge.frequency = 1$
					}}
				}
			}
		}
		$currentMSLevel = currentMSLevel+1$
	}}
	\caption{Learning Algorithm for SAGE. $currentMSLevel$ represents the current MS level in the MS data we are processing, we start with level $0$ which is related to MS level 1. $maxMSLevel$ is the highest level in the given MS data.}
\end{algorithm}

\begin{figure}[!tpb]
	\centerline{\includegraphics[scale=0.4]{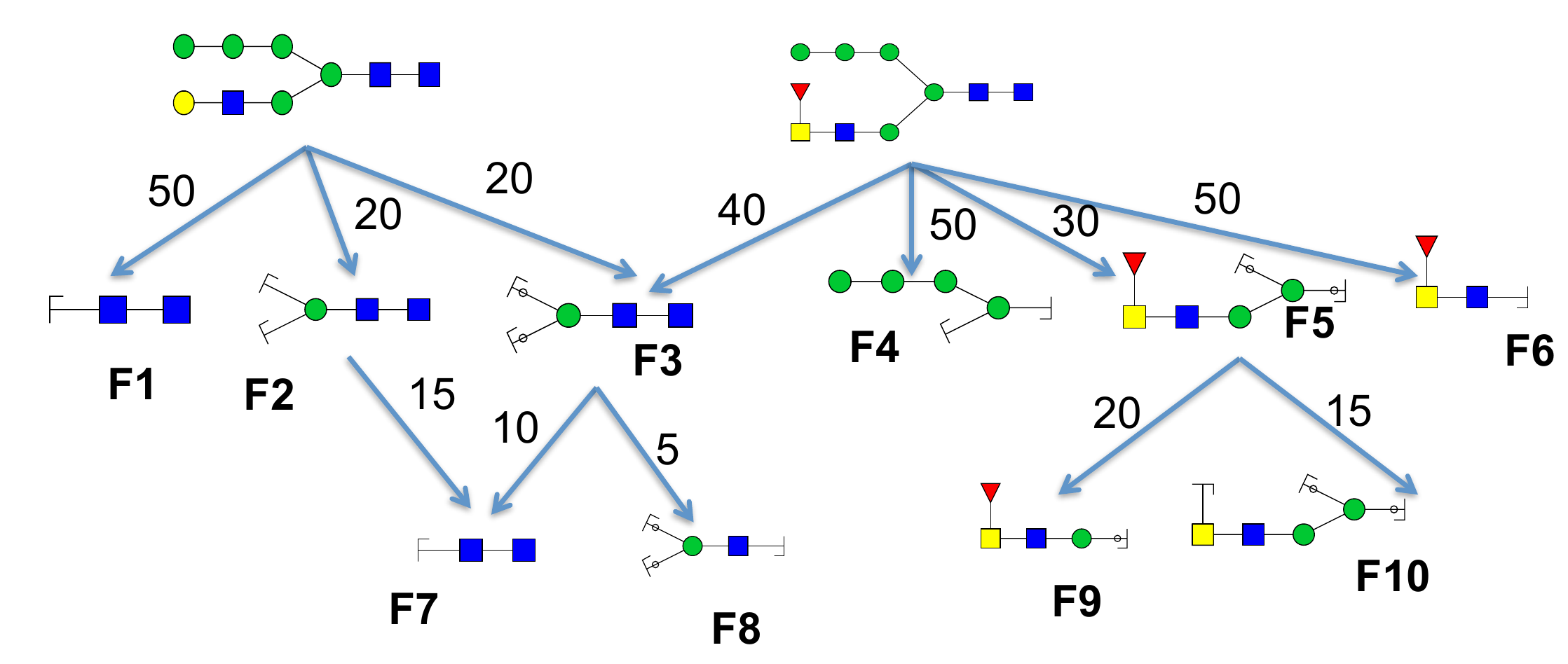}}
	\caption{SAGE representing the MS data up to MS$^3$. The root nodes are glycans used to annotated precursors in the training data, while the nodes at lower levels represent fragments (F$_i$ is the fragment Id) used to annotate the peaks in different MS levels. The edges represent the co-occurrence between the two nodes it connect while the number on the edge represent the frequency of that co-occurrence.}\label{fig:01}
\end{figure}

SAGE approaches MS annotation as a multi-label classification problem. Complete glycan structures that can annotate the spectra are the classes (into which observed spectra are assigned) while the fragments used to annotate the observed MS peaks are treated as features. Figure \ref{fig:01} illustrates the representation of MS data in SAGE. Each root node is labeled with a glycan structure and a specific $m/z$ value, indicating that the user approved that structure during the training phase to annotate a precursor ion observed at that $m/z$.
The nodes in the lower levels represent the fragments approved to annotate $MS^2$, $MS^3$, $MS^n$ peaks. Edges are allowed between nodes at level i to nodes at level $i + 1$ if and only if the ion observed and selected at level $i$ decomposes to generate the fragment ion at level $i + 1$. Numeric edge labels represent the number of times the parent node (the source of the edge) appears in the training data in association with the child node (the destination of the edge).
After training,  SAGE processes new (unannotated) data by associating the precursor ion of each MS scan with a set of root nodes (classes) and associating $m/z$ values from the scan (peak list) with features in the trained model. The probabilistic classification score, $P(G_x | f_1, f_2, f_3, f_n)$ of each glycan (root node) is calculated given observation of those features in the scan. In order to optimize the search space, only those glycans forming quasi-molecular ions with $m/z$ values that are within a specified tolerance from the precursor $m/z$ for the given scan are considered. Figure \ref{fig:annotation_workflow} shows the annotation workflow for SAGE. The probabilistic score is calculated as described in \cite{aljadda2014pgmhd}. For example, to use SAGE instance in Figure \ref{sage}, assume we have a spectra with the fragments $F_1,F_3,F_7$ and we would like to know which glycan structure in the root level is the best annotation. We can calculate the probabilistic score $P(G_1|F_1,F_3,F_7)$ and $P(G_2|F_1,F_3,F_7)$ as described in \cite{aljadda2014pgmhd}. 
\begin{equation*}
\begin{split}
P(G_1|F_1,F_3,F_7) &= P(G_1|F_1,F_3)*P(F_3|F_7)\\
&= P(G_1|F_1)*P(G1|F_3)*(F_3|F_7)\\
&= 50/50 * 20/60 * 10/25 = 0.13
\end{split}
\end{equation*}
\begin{equation*}
\begin{split}
P(G_2|F_1,F_3,F_7) &= P(G_2|F_1,F_3)*P(F_3|F_7)\\ 
&= P(G_2|F_1)*P(G2|F_3)*(F_3|F_7)\\ 
&= 0.1 * 40/60 * 10/25 = 0.02
\end{split}
\end{equation*}

Since there is no edge between $G_2$ and $F_1$ even though it is a valid and possible fragment to $G_2$, to resolve the zero probability problem. Any fragment that cannot possibly be generated by fragmentation of a candidate glycan will not be included due to the selectivity of the model.

\begin{table*}[t]
	\newcolumntype{L}[1]{>{\raggedright\let\newline\\\arraybackslash\hspace{0pt}}m{#1}}
	\newcolumntype{C}[1]{>{\centering\let\newline\\\arraybackslash\hspace{0pt}}m{#1}}
	\newcolumntype{R}[1]{>{\raggedleft\let\newline\\\arraybackslash\hspace{0pt}}m{#1}}
	
	\centering
	\caption{Features of the existing MS annotation tools. Annotates MS$^n$, reflects the ability of the tools to annotate MS$^n$ spectra, not just MS profile. Annotates with structures, means the tool uses database of glycans for annotation. Can find novel structures, means it can annotate using new glycan structures it never used before for annotation. Scores, define what type of scores the tools support. Statistical means scores calculated using statistical methods (summation, counting,etc.), while probabilistic means scores calculated based on probability distribution. Handle under-methylation, reflects the ability of the tools to consider under-methylation during the annotation process. Handle natural loss, reflects the ability to consider the natural loss in the annotation process. Average annotation time, reflects in general how long the annotation process require to be done. And availability, is either the tool is freely available or it is a commercial one.}
	\label{tbl:comp}
	\begin{adjustwidth}{-1.4in}{-1.1in}
	\begin{tabularx}{1.1\linewidth}{|X|X|X|X|X|X|X|X|X|X|}
		\hline
		Tool & Annotates MS$^n$ & Annotates with structures & Can find novel structures & Scores & Handle Under-methylation & Handle Natural Loss & Uses human expert knowledge & Avg Annotation Time & Availability\\
		\hline
		GELATO & Yes & Yes & Yes & Statistical & Yes & Yes & No & Minutes & Free\\
		\hline
		SAGE & Yes & Yes & No & Probabilistic & No & No & Yes & Seconds & Free\\
		\hline
		SAGE and GELATO & Yes & Yes & Yes & Statistical and Probabilistic & Yes & Yes & Yes & Minutes & Free\\
		\hline
		GWB & Yes & Yes & Yes & Statistical & No & No & No & Minutes & Free\\
		\hline
		GlycoMod & No & Yes & Yes & Statistical & No & No & No & Minutes & Free \\
		\hline
		SimGlycan & Yes & Yes & Yes & Statistical & No & No & No & Hours & Commercial \\
		\hline
	\end{tabularx}
\end{adjustwidth}

\end{table*}


\section{Discussion}
\begin{figure}[!tpb]
	\centerline{\includegraphics[scale=0.35]{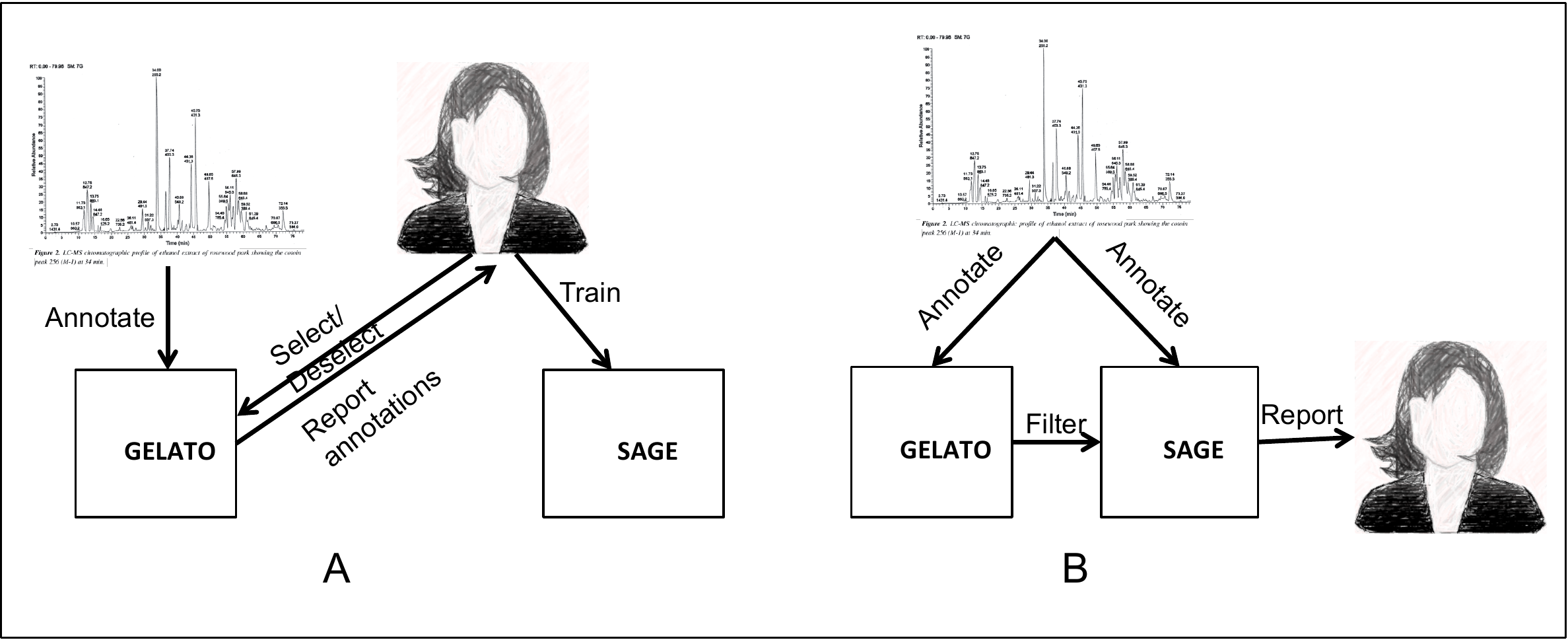}}
	\caption{The integration between SAGE and GELATO. (A) GELATO annotates a given spectra, then a user select subset of those annotations and provide this final list of approved annotations to train SAGE. (B) the trained SAGE can be used either to annotate the given spectra or to filter out the annotations calculated by GELATO which most likely will not be selected by the user.}\label{fig:02}
\end{figure}

\begin{figure}[!tpb]
	\centerline{\includegraphics[scale=0.35]{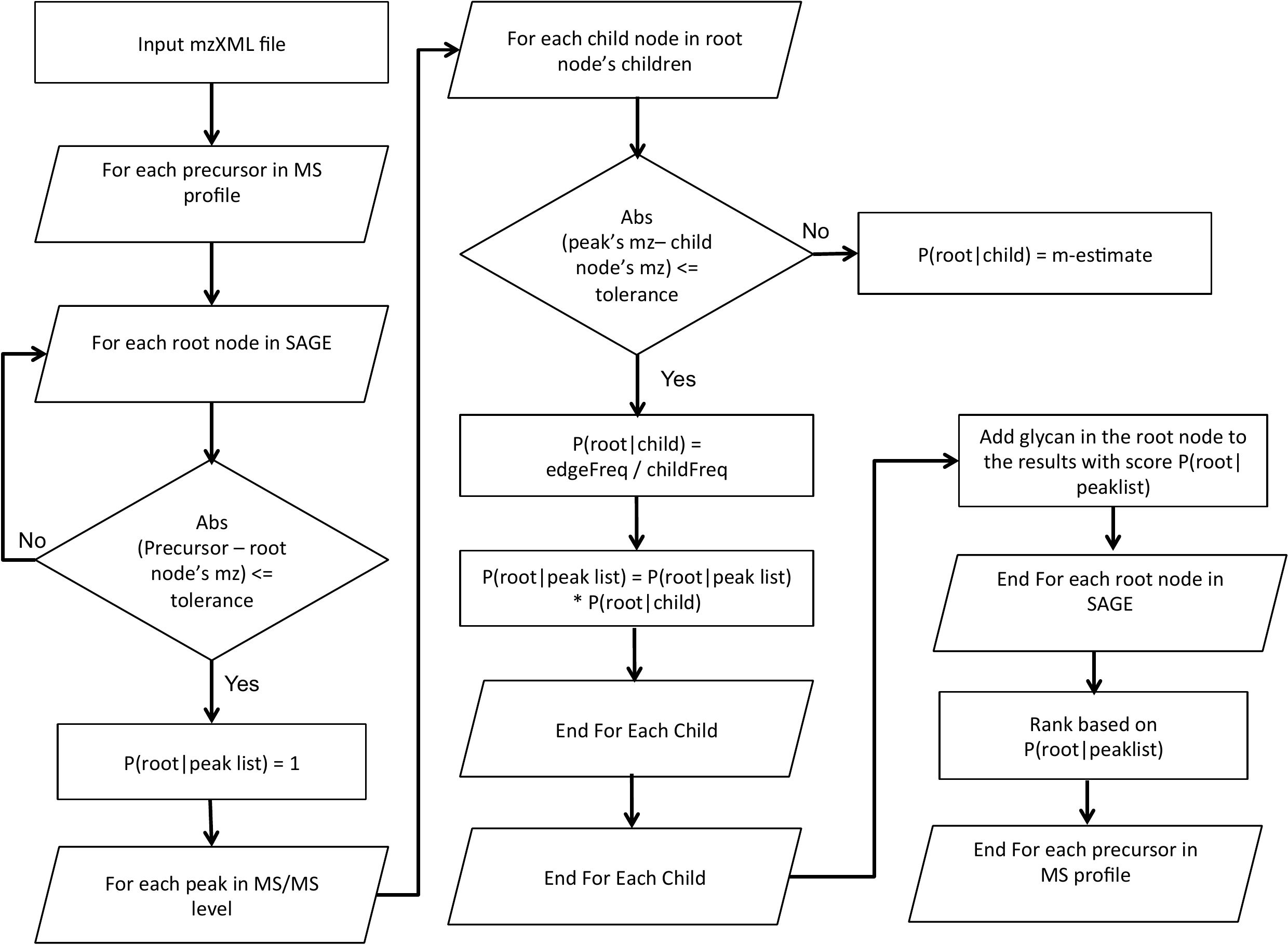}}
	\caption{The annotation workflow of SAGE. For more information about m-estimate, childFreq, edgeFreq and probabilistic based score please see \cite{aljadda2014pgmhd}}\label{fig:annotation_workflow}
\end{figure}

The software described here combines the semi-automated MS annotation tool GELATO with the machine learning model SAGE to automate the process of MS annotation.  As shown in Figure \ref{fig:02} A, the training of SAGE depends on the annotations generated by GELATO. The human analyst then examines and evaluates these annotations to select a subset that are deemed accurate and relevant. The selected annotations are then used to train SAGE. Figure \ref{fig:02} B, illustrates application of the trained SAGE for the de novo annotation of new spectra or improvement of the annotations previously generated by GELATO. This post-filtering process uses the knowledge learned by SAGE to eliminate the annotations that are most likely to be rejected by the user.
SAGE thus calculates a probabilistic score for each possible annotation (scan-glycan pairing), using previously learned knowledge of the annotation choices made by an analyst or group of analysts and reflecting the likelihood that the annotation would be accepted. The user can instruct SAGE to report the k top-scoring annotations. Otherwise, SAGE will report all the possible annotations ranked by their probabilistic score. This integrated framework addresses many of the shortcomings of existing tools, which are listed in Table \ref{tbl:comp}.

As human knowledge is subjective, annotations that seem correct to one person may not be judged as correct by another person.  Thus, SAGE not intended to provide a global knowledgebase that can be applied everywhere by everyone. Each instance of SAGE represents a particular analyst’s knowledge, which he/she would like to apply to the annotation of new MS data sets. In this context, it is important to train new instances of SAGE to annotate MS data generated by analysis of different types of glycans (e.g., N-glycans vs. O-glycans) and the glycans from different taxonomic species, each of which are capable of generating their own unique collection of glycan structures.


\section{Experiment and Results}
\begin{figure}[!tpb]
	\centerline{\includegraphics[scale=0.5]{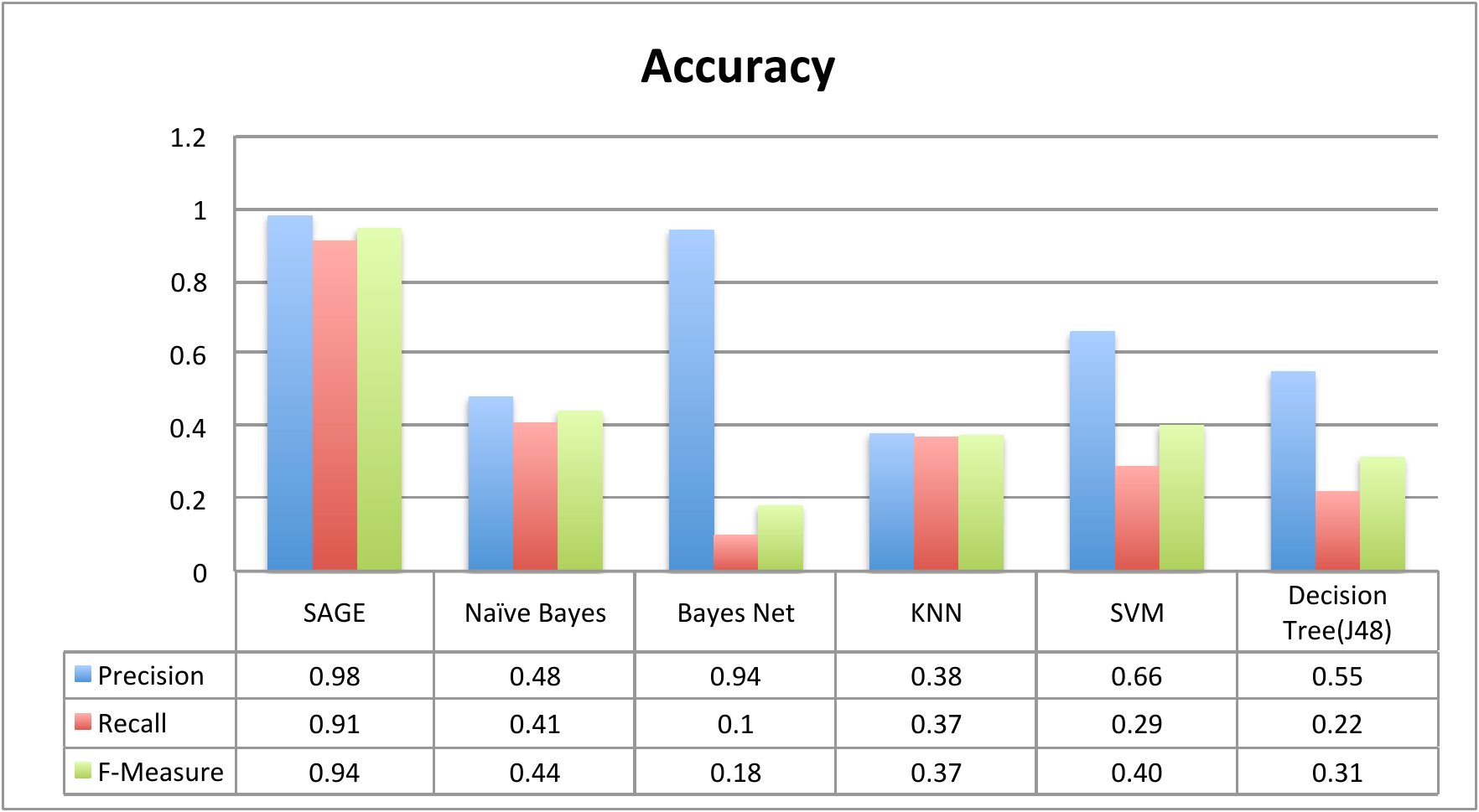}}
	\caption{Precision, Recall, and F-Measure of the SAGE compared to the most popular classifiers}\label{fig:04}
\end{figure}

\begin{figure}[!tpb]
	\centerline{\includegraphics[scale=0.50]{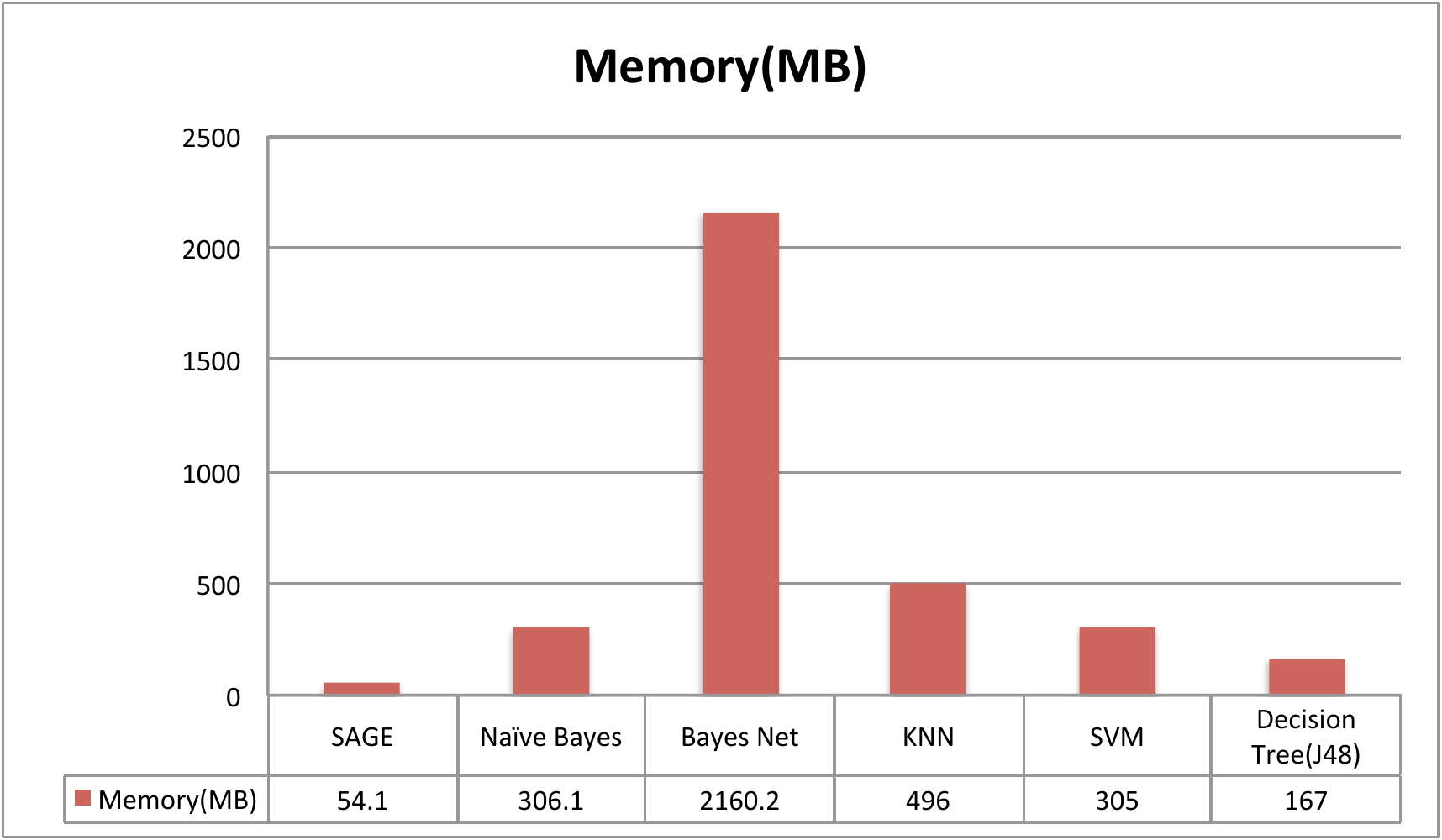}}
	\caption{Main memory usage by SAGE compared to the other machine learning models for the training dataset}\label{fig:05}
\end{figure}

To test our data-processing framework, we used GELATO to annotate 10 MS datasets generated during the glycoanalysis of stem cells. Annotations generated by GELATO were compared with annotations generated by SimGlycan, a commercial software tool specifically designed for the this purpose. For each spectrum tested, GELATO generated all the annotations generated by SimGlycan. However, GELATO was able to provide additional annotations that SimGlycan could not due to the following limitations.  Firstly, SimGlycan cannot annotate scans whose precursor ions are not previously assigned a definite charge state (which is indicated in the input mzXML file). In contrast, GELATO identifies and annotates precursor ions with multiple charges by evaluating all charge states between 1 and the maximum charge (inclusive) specified by the user. Secondly, the database provided by and used by SimGlycan is incomplete. However, GELATO uses a custom glycan database generated from a highly curated source, such as the GlycO ontology. 
The integrated annotation framework was tested in a second experiment, where GELATO was used to annotate MS data obtained by glycoanalysis of 10 different samples from pancreatic cancer patient \cite{porterfield2013discrimination}. An MS expert reviewed the annotations generated by GELATO and selected a subset that she judged to be correct. Nine of the ten curated annotations were used to train SAGE, while the tenth was used for testing. The trained SAGE was used to generate de novo annotations of the tenth MS data set and these annotations were compared to those generated by GELATO and approved by the expert for the same dataset. This process was repeated 10 times with different test sets and training sets, and the average accuracy of the annotations was calculated. By these criteria, the annotations generated by SAGE were, on average, $98\%$ accurate with $91\%$ coverage (SAGE predicted $91\%$ of the manually approved annotations of the test spectra). 
Since SAGE functions as a multi-label classifier and the MS annotation is multilabel classification problem, we compared SAGE with the top classifiers in machine learning. We used Mulan \cite{tsoumakas2011mulan} which is an extension to the well-known machine learning library Weka to handle multi-label classification. Figure \ref{fig:04} shows that the SAGE implementation of PGMHD outperforms all the well-known classifiers in this challenging task. We are integrating our glycoanalysis annotation framework into the GRITS-Toolbox (http://www.grits-toolbox.org), a standalone application for the interpretation and annotation of glycomics MS data. Given that the GRITS-Toolbox is designed to run on a single workstation (desktop or laptop) with limited memory, control of memory usage is a critical issue. In this context, we compared the memory usage of SAGE compared to the other classifiers. Figure \ref{fig:05} shows that the SAGE implementation of PGMHD used less memory (54 megabytes on average) than the other machine learning models to represent the training dataset. Another critical aspect of the training of and the annotation by SAGE is its time complexity. Figure 4.9 shows that, of all the tested machine learning models, SAGE required the least training time when using the same training dataset. Figure \ref{fig:07} compares the annotation time using different machine learning models,showing that SAGE is the third fastest in this respect.

\begin{figure}[!tpb]
	\centerline{\includegraphics[scale=0.5]{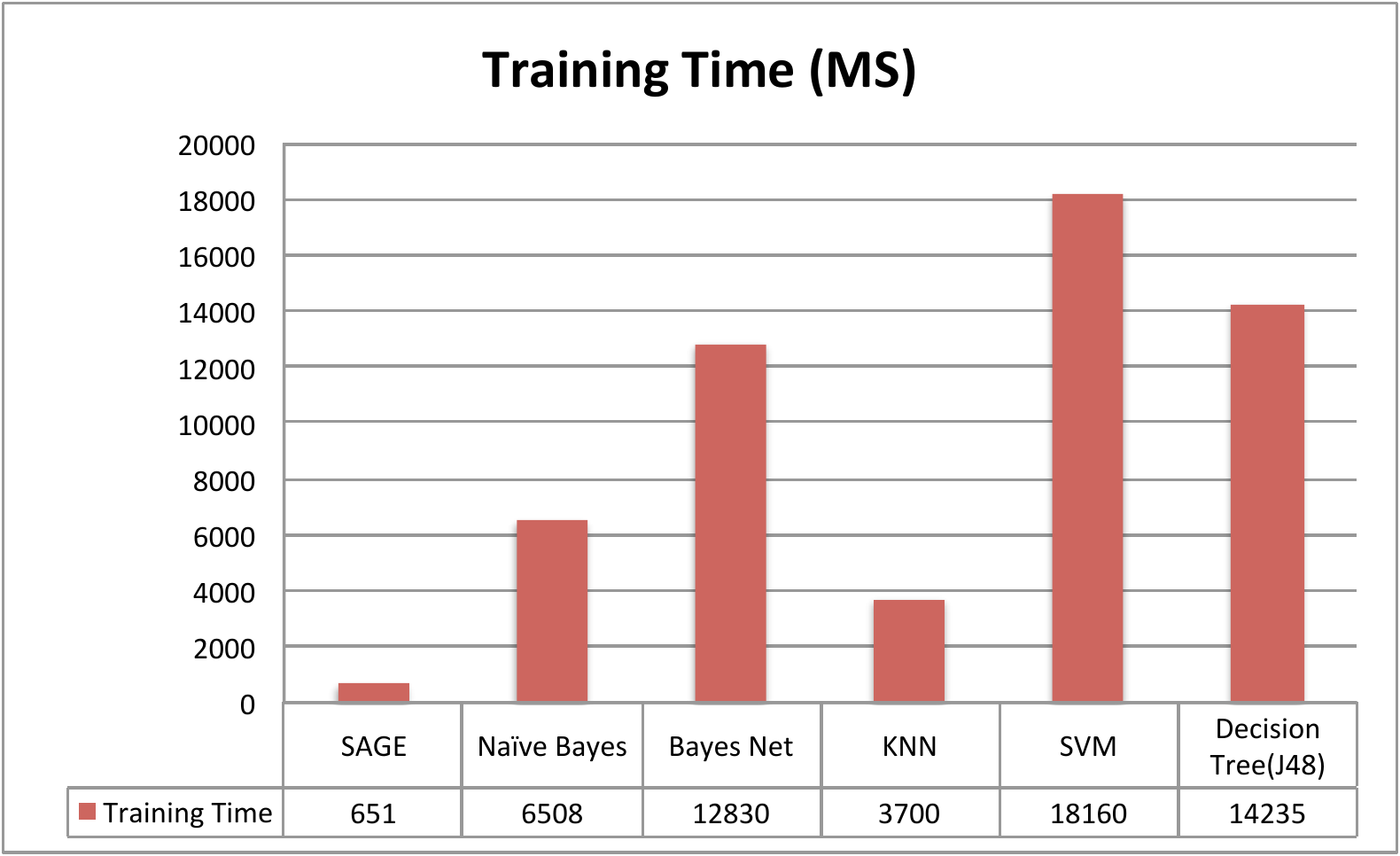}}
	\caption{Training time for different models in millisecond. SAGE is the fastest model to learn and converge.}\label{fig:06}
\end{figure}

\begin{figure}[!tpb]
	\centerline{\includegraphics[scale=0.5]{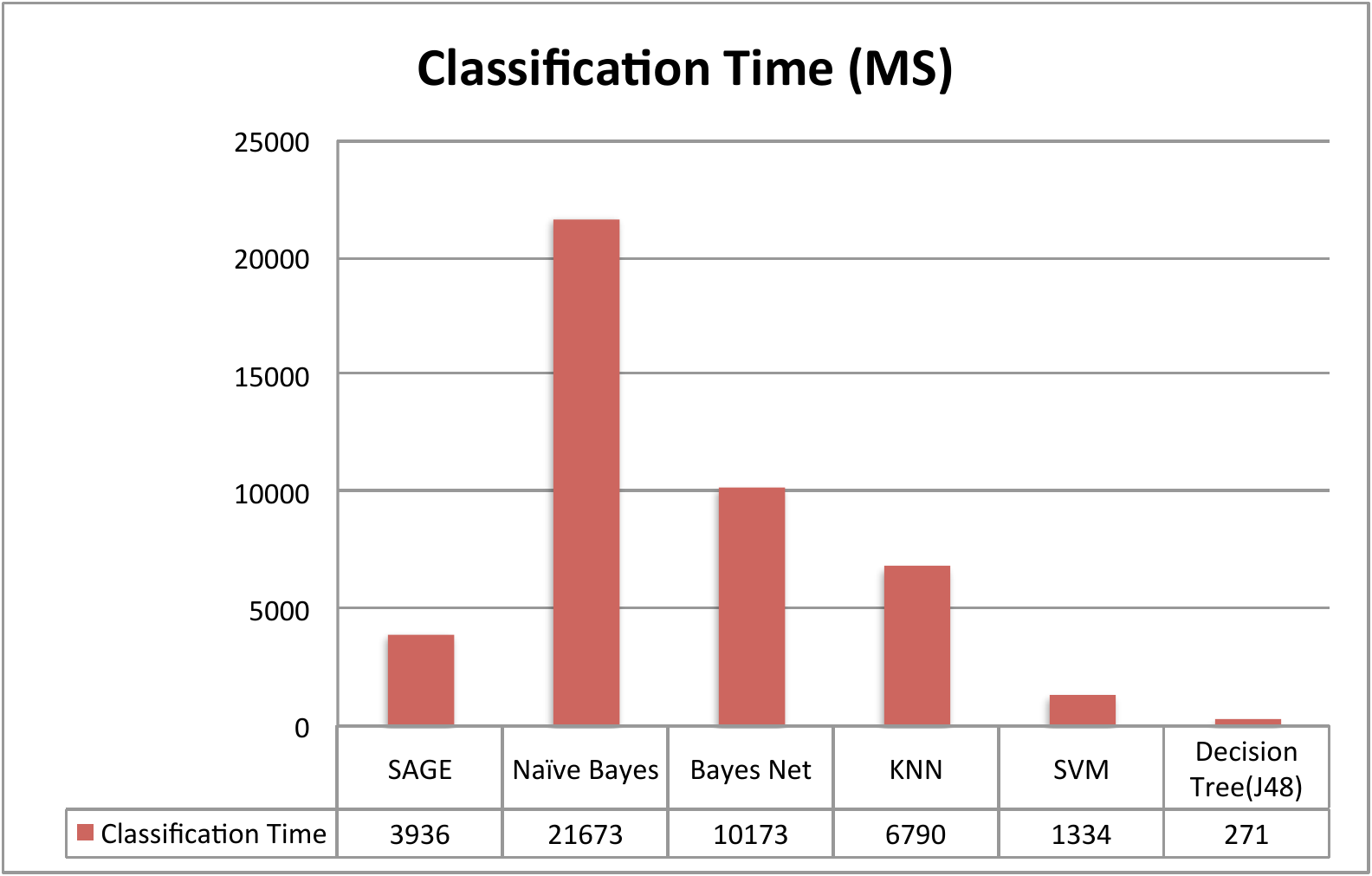}}
	\caption{Annotation time for different models in millisecond. SAGE is the third fastest model.}\label{fig:07}
\end{figure}

%
%

\section{Conclusion}
We have designed and implemented an integrated software package for automated annotation MS data with glycan structures, This package consists  of two major components, GELATO and SAGE. GELATO is a semi-automated annotation tool which is built upon two open source projects GlycoWorkbench and GlycomeDB. SAGE is a novel machine learning model that mimics the annotation patterns of an expert in MS interpretation to identify annotations of new data that are likely to be accepted or rejected by a human analyst . The current implementation of this framework utilizes GELATO to generate annotations that are used evaluated by a human.  The selected annotations to train SAGE, which can then be used for either the de novo annotation of MS spectra or the improvement of annotations generated by GELATO.  SAGE provides a probabilistic score that can be used as the basis for rejecting annotations that are likely to be incorrect.

\section*{Acknowledgement}
We would like to thank Camilo Ortiz, and Trey Grainger from Careerbuilder.com for their contribution to the PGMHD model which we customize to SAGE. Also, we would like to thank Kiyoko Aoki Kinoshita from Soka University for the valuable time and deep discussion with her during the design of GELATO and SAGE. Many thanks to the team of EUROCarbDB for the well designed open source libraries which we built GELATO upon.
\paragraph{Funding} 
This work was supported by the National Institute of General Medical Sciences, a part of the National Institutes of Health, funding the National Center for Biomedical Glycomics (8P41GM103490).

\bibliographystyle{plain}
\bibliography{main}

\end{document}